\pdfoutput=1

\documentclass[11pt]{article}

\usepackage[preprint]{acl}

\usepackage{times}
\usepackage{latexsym}

\usepackage[T1]{fontenc}

\usepackage[utf8]{inputenc}

\usepackage{microtype}

\usepackage{inconsolata}

\usepackage{graphicx}
\usepackage{amsmath}
\usepackage{multirow}
\usepackage{booktabs}
\usepackage{amsfonts}

\title{MA-COIR: Leveraging Semantic Search Index and Generative Models for Ontology-Driven Biomedical Concept Recognition}

\author{
 \textbf{Shanshan Liu\textsuperscript{1,2}},
 \textbf{Noriki Nishida\textsuperscript{1}},
 \textbf{Rumana Ferdous Munne\textsuperscript{1}},
 \textbf{Narumi Tokunaga\textsuperscript{1}},
\\
 \textbf{Yuki Yamagata\textsuperscript{3, 4}},
 \textbf{Kouji Kozaki\textsuperscript{5}},
 \textbf{Yuji Matsumoto\textsuperscript{1}},
 \\
    \textsuperscript{1}RIKEN AIP \ \ \ 
    \textsuperscript{2}University of Tsukuba \ \ \
    \textsuperscript{3}RIKEN R-IH \ \ \
    \textsuperscript{4}RIKEN BRC \ \ \ \\
    \textsuperscript{5}Osaka Electro-Communication University\\
    \{shanshan.liu, noriki.nishida, rumanaferdous.munne, narumi.tokunaga, \\
    yuki.yamagata, yuji.matsumoto\}@riken.jp\\
    kozaki@osakac.ac.jp\\
}

\begin{document}
\maketitle

\begin{abstract}

Recognizing biomedical concepts in the text is vital for ontology refinement, knowledge graph construction, and concept relationship discovery. However, traditional concept recognition methods, relying on explicit mention identification, often fail to capture complex concepts not explicitly stated in the text. To overcome this limitation, we introduce MA-COIR, a framework that reformulates concept recognition as an indexing-recognition task. By assigning semantic search indexes (ssIDs) to concepts, MA-COIR resolves ambiguities in ontology entries and enhances recognition efficiency. Using a pretrained BART-based model fine-tuned on small datasets, our approach reduces computational requirements to facilitate adoption by domain experts. Furthermore, we incorporate large language model (LLM)-generated queries and synthetic data to improve recognition in low-resource settings. Experimental results on three scenarios (CDR, HPO, and HOIP) highlight the effectiveness of MA-COIR in recognizing both explicit and implicit concepts without the need for mention-level annotations during inference, advancing ontology-driven concept recognition in biomedical domain applications. Our code and constructed data is available at \url{https://github.com/sl-633/macoir-master}.
\end{abstract}

\begin{figure}[t]
\centering
  \includegraphics[width=0.9\linewidth]{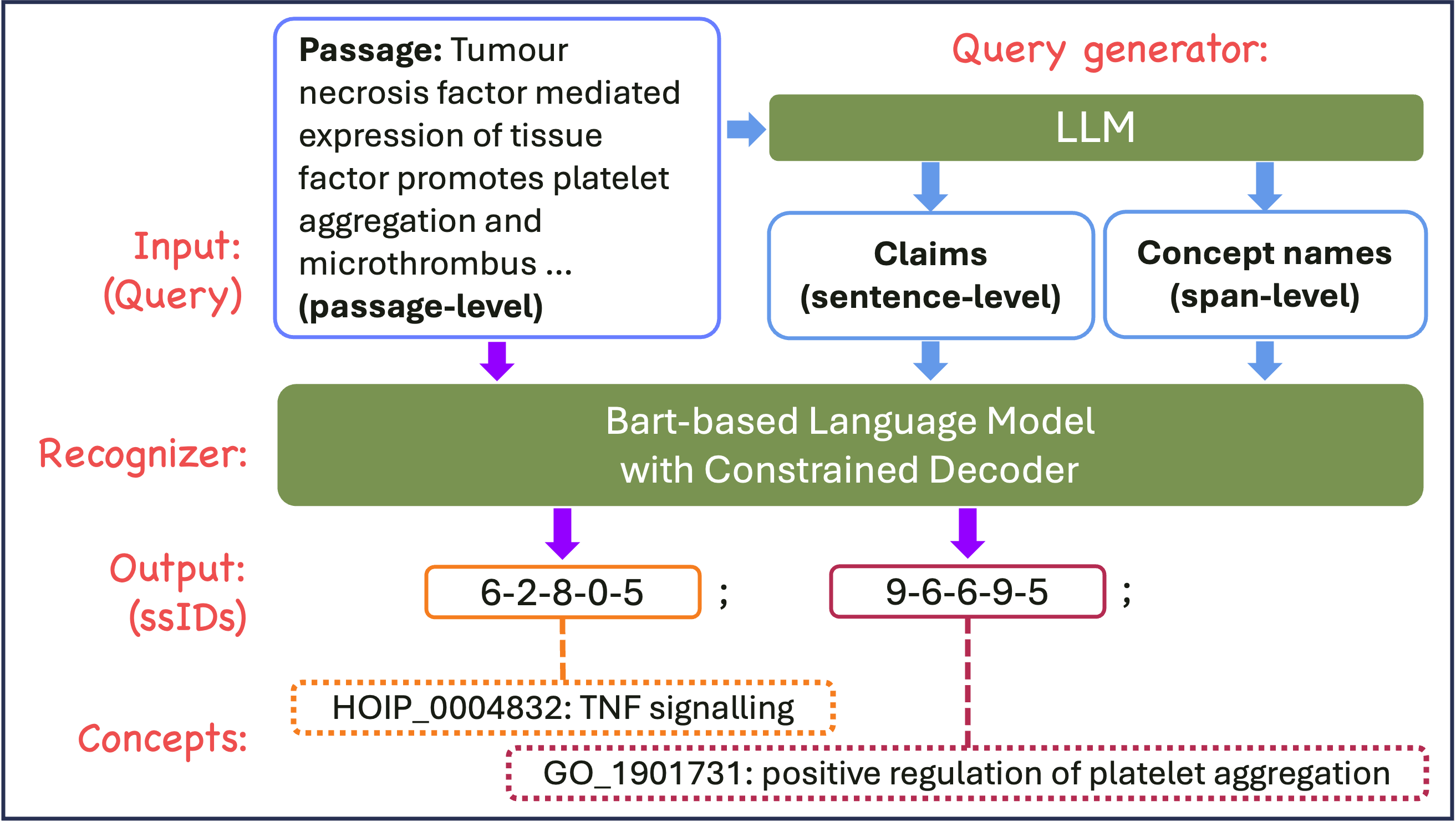}
  \caption {Concept recognition by MA-COIR follows the default workflow indicated by purple arrows. When an LLM generates simplified queries from a given passage, additional processes, denoted by blue arrows, are incorporated. When ``6-2-8-0-5'' is generated, ``HOIP\_0004832: TNF signalling'' is predicted as a concept within the query.}
  \label{fig:overview-workflow}
\end{figure}

\section{Introduction}

Automatic recognition of biological concepts in the text aids experts in refining ontologies and consolidating domain knowledge. As structured knowledge evolves to include increasingly complex concepts \cite{10.1093/nar/gkad1005,Yamagata2024}, identifying concepts often requires significant expert analysis. Traditional Concept Recognition (CR) methods are inadequate for supporting tasks such as ontology-driven knowledge graph construction, efficient literature retrieval for specific concepts, and the discovery of novel relationships between concepts.

Typically, recognizing ontology concepts in passages or sentences relies on identifying mentions - text spans where concepts appear. When mentions are provided, Entity Disambiguation (ED) can be applied to match each mention to a single entity or none at all \cite{wu2019zero,jiang2024entity,wang2023exploringincontextlearningability,oaklib}. When mentions are unknown, recognition may be achieved through a pipeline beginning with Named Entity Recognition (NER) to identify mentions, followed by ED to resolve these predictions \cite{shlyk-etal-2024-real,10.1093/bioinformatics/btae104}. Alternatively, end-to-end Entity Linking (EL) approaches can yield a series of (mention, entity) pairs \cite{kolitsas-etal-2018-end,DBLP:journals/corr/abs-2010-00904,luo2021phenotagger}.


With advancements in Large Language Models (LLMs), several LLM-based pipeline methods for NER and ED have been introduced \cite{shlyk-etal-2024-real,10.1093/bioinformatics/btae104}. In-context learning (ICL) techniques reduce annotation requirements; however, a substantial performance gap remains between ICL and fully supervised methods \cite{shlyk-etal-2024-real}. While mention-based queries are typically generated to retrieve concepts, the limitation of this approach becomes evident when complex concepts do not appear explicitly as ``mentions'' within the text, rendering aforementioned mention-based recognition methods ineffective in real-world applications.


We propose \textbf{MA-COIR} (\textbf{M}ention-\textbf{A}gnostic \textbf{Co}ncept Recognition through an \textbf{I}ndexing-\textbf{R}ecognition Framework), a framework for recognizing biomedical concepts explicitly or implicitly mentioned in the text. Inspired by prior works~\cite{tay2022transformermemorydifferentiablesearch,jiang2024entity}, we reformulate the concept recognition (CR) task into an indexing-recognition paradigm. This approach assigns each concept a semantic search index (ssID) and trains a neural model to predict ssIDs corresponding to concepts described in the input text (see Fig. \ref{fig:overview-workflow}).

By generating ssIDs instead of literal concept names, the framework resolves ambiguities caused by identical concept names within ontologies (e.g., concepts sharing preferred names but differing definitions). Additionally, the semantic alignment between concepts and their assigned indexes enhances model learning, enabling more efficient recognition.

Our method leverages a pretrained BART-based language model fine-tuned on a small dataset, thereby reducing computational demands and improving accessibility for domain experts. Furthermore, we explore LLM-generated queries and synthetic data, demonstrating the framework's utility in low-resource settings for real-world concept extraction tasks.

Results across datasets (CDR, HPO, and HOIP) demonstrate the effectiveness of our framework.

Our contributions are:
\begin{itemize}
    \item We propose MA-COIR, a novel framework for recognizing both explicit and implicit biomedical concepts without the need for prior identification of specific mentions, thereby reducing reliance on labor-intensive annotations needed for entity recognizer training. 
    \item To the best of our knowledge, we are the first to integrate a semantic search index into concept recognition, improving generative model learning and enabling more efficient recognition. 
    \item We demonstrate the utility of query and training data generated by an LLM in concept recognition tasks, providing a reference framework for efficient recognition in low-resource settings.
\end{itemize}





\section{Related work}

In recent years, biomedical CR methods have largely followed two main approaches. The first approach involves fully-, weakly-, or self-supervised learning methods based on pretrained language models, such as domain-specific BERT or BART models \cite{liu-etal-2021-self,10.1093/bioinformatics/btz682,yuan-etal-2022-biobart,zhang-etal-2022-knowledge}, and fine-tuned these models on small annotated datasets \cite{luo2021phenotagger}. The second approach leverages the strong generalization capabilities of LLMs to perform NER and ED tasks in zero- or few-shot settings \cite{wang2023exploringincontextlearningability}.

Existing biomedical CR methods that operate without mention annotations are LLM-based. For instance, \cite{10.1093/bioinformatics/btae104} explored a schema guiding LLMs to perform NER with specified constraints, using \cite{oaklib} for subsequent ED tasks. \cite{shlyk-etal-2024-real}  proposed the REAL framework, which combines LLM-based zero-shot NER with an ED method using retrieval-augmented generation (RAG). \cite{el-khettari-etal-2024-mention} developed an ICL demonstration selection strategy to generate concept names closely aligned with ontology terms, subsequently linking them based on the similarity between generated names and ontology terms.




\section{Methodology}

\subsection{Task formulation}

Let \(O\) represent a set of concepts \(\{C_1, ..., C_n\}\) defined within a domain ontology. Given a query text \(Q\), the CR task aims to identify a subset of concepts \(\{C'_1, ..., C'_p\}\) from the ontology that are referenced in the text.

We approach the CR task as an end-to-end generative process. First, we assign each concept \(C\) a unique semantic search index (ssID). Then, our model generates one or more ssIDs for the input text \(Q\), thereby retrieving the concepts are presented in the text.

\subsection{Concept Indexing}
\label{sec:concept-indexing}

\begin{figure}[t]
\centering
  \includegraphics[width=0.75\linewidth]{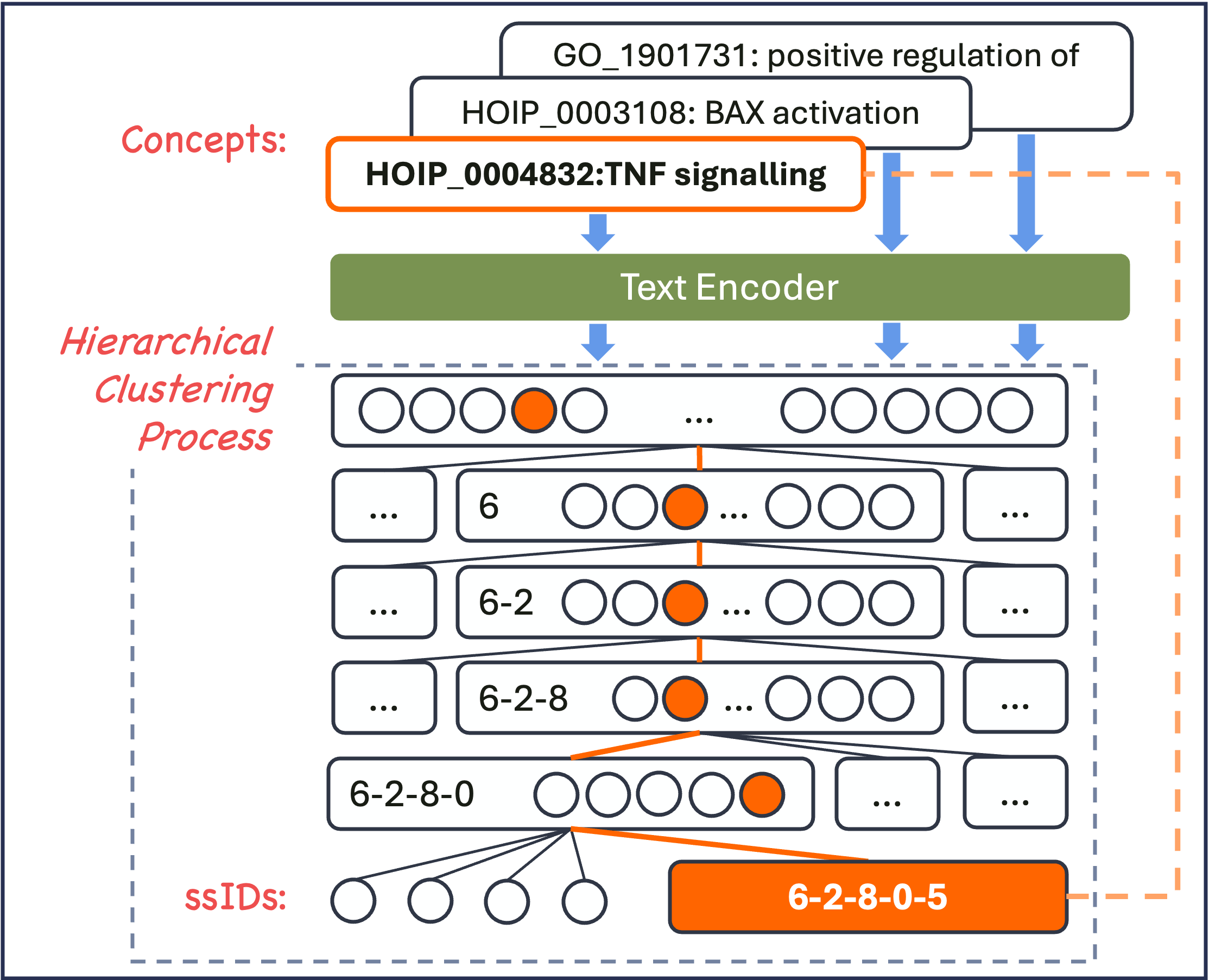}
  \caption {Indexing Phase in MA-COIR: A semantic search index (ssID) is assigned based on a label tree derived from the domain ontology. Through hierarchical clustering, the ssID for the concept ``HOIP\_0004832: TNF signaling'' is ``6-2-8-0-5''.}
  \label{fig:indexing}
\end{figure}


As illustrated in Fig. \ref{fig:indexing}, each concept \(C\) is represented as a vector \(E_C\), obtained by encoding its canonical name \(Name_C\) using a text encoder. Given our focus on the biomedical domain, we select SapBERT \cite{liu-etal-2021-self} as the text encoder.\footnote{Through preliminary experiments, we observed that using the average of token embeddings yields better performance than the [CLS] token. We evaluated several pretrained language models, including BioBERT v1.1, PubMedBERT, SapBERT, and SciBERT, with SapBERT achieving the best results.} The representation \(E_C\) is derived by averaging the last hidden states for the tokens in \(Name_C\).
\begin{equation}
  \label{eq:1}
  X_C = TextEncoder(Name_C) \in \mathbb{R}^{l\times H}
\end{equation}
\begin{equation}
  \label{eq:2}
  E_C = avg(X_C) \in \mathbb{R}^H
\end{equation}
where \(l\) is the token length, and \(H\) is the dimension of each token's embedding.

Starting with the ROOT node that encompasses all concepts in the target ontology, we construct a label tree using a top-down \textbf{hierarchical clustering process}. Specifically, if a node contains more than \(g\) elements, we divide it into \(\leq m\) categories until each leaf node corresponds to a single concept (with \(g=10, m=10\) in this study) by K-means algorithm implemented with Scikit-learn \cite{scikit-learn}. Each node is assigned an index based on its category, forming a sequence of ``semantic search indexes'' (ssIDs) that encode semantic information from each concept's representation.

\begin{figure}[t]
    \centering
    \includegraphics[width=0.95\linewidth]{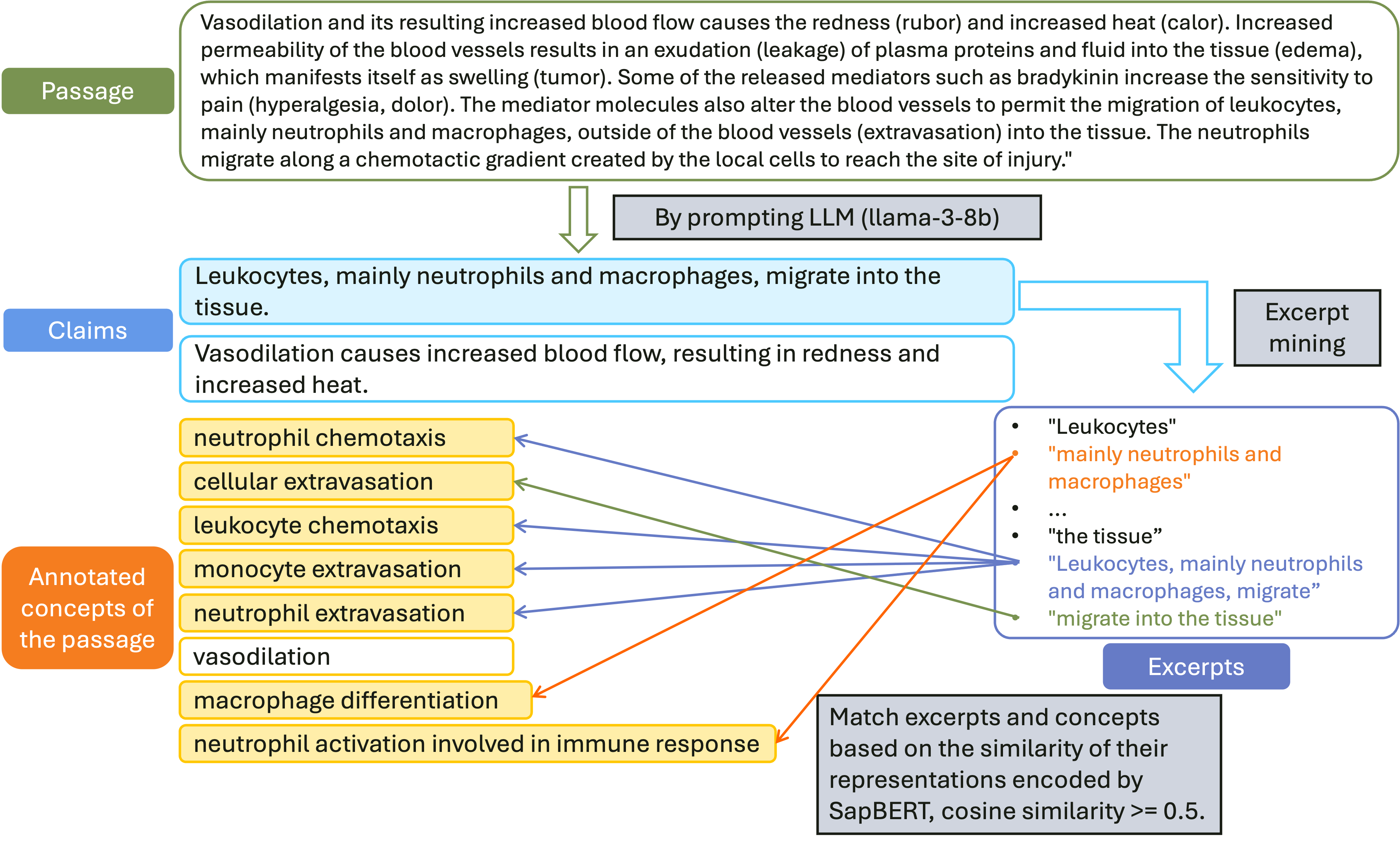}
    \caption{An example of constructing a claim-concept instance is as follows: Given a passage, 
    we prompt the LLM to breakdown the passage into several claims. For \textcolor{blue}{one claim}, we then perform excerpt mining. 
    Next, we match these mined excerpts to the passage’s annotated concepts by assessing semantic similarity. If an excerpt closely aligns with an annotated concept, we pair the concept with the claim. In this example, \textcolor{orange}{seven concepts} are paired with a single claim, forming a claim-concept instance.}
    \label{fig:claim-concept}
\end{figure}

\subsection{Concept Recognition}
\label{sec:concept-recognition}
During recognition phase following the indexing process, the input may consist of a passage (e.g., a paragraph of one PubMed article), a sentence, or a span (mention or concept name), while the output is a text sequence listing ssIDs (e.g., ``6-2-8-0-5; 9-6-6-9-5;''). Each ssID is separated by a semicolon (``;''), as illustrated in Fig. \ref{fig:overview-workflow}.

To effectively map natural language text to a formatted sequence, we selected a BART-based pretrained language model (facebook/bart-large) \cite{DBLP:journals/corr/abs-1910-13461}. This model, with its encoder-decoder architecture and cross-attention mechanism, is well-suited for our tasks.

To ensure the BART-based model generates valid ssID sequences, we apply a constrained decoder that filters the output to retain only valid ssIDs. The decoder's vocabulary \(T\) is restricted to ssID tokens. The token embedding \(e_t\) for each token \(t \subset T\) is obtained from the embedding layer \(LmEmbedding\) of the language model \(LM\):

\begin{equation}
  \label{eq:3}
  e_t = LmEmbedding(t) \in \mathbb{R}^{H}
\end{equation}
where \(H\) is the dimension of a token's embedding.

At the \(i\mathchar`-{th}\) time step, the decoder selects the token with the highest score based on the token embedding \(e_t\) and the last hidden state \(h_i\). One feature \(h_{i, t}\) is computed using a one-layer linear classifier:
\begin{equation}
  \label{eq:4}
  h_{i} = LM(\hat{y}_{i-1}) \in \mathbb{R}^{H}
\end{equation}
\begin{equation}
  \label{eq:5}
  h_{i, t} = W^{o}_t h_{i} + b^o
\end{equation}
where \(W^{o}\) is the weight and \(b^{o}\) is the bias of the classifier.

Another feature \(e_{i, t}\) is the dot product of \(e_t\) and \(h_i\), representing the relevance between the token \(t\) and \(h_i\):
\begin{equation}
  \label{eq:6}
  e_{i, t} = e_t h_{i} 
\end{equation}
The final score of the token \(t\) is the average of two features:
\begin{equation}
  \label{eq:7}
  z_{i, t} = avg(e_{i, t}, h_{i, t}))
\end{equation}
\begin{equation}
  \label{eq:8}
  \hat{y}_i = \arg\max_{t}(\sigma(z_{i,t}))
\end{equation}
where \(h_{i, t}, e_{i, t}, z_{i, t} \in \mathbb{R}^{1}\), \(\sigma\) is the Softmax function. The model parameters are optimized by minimizing the \(CrossEntropyLoss(y, \hat{y})\).

Our preliminary experiments revealed that using only one canonical name-ssID pair to introduce a concept into the model did not provide strong performance. It is crucial to incorporate synonym-, mention-, and passage-ssID pairs for model improvement if they are available. Therefore, our model is trained on various input-output pairs. When the input is a span and the output is the ssID of a single concept, the model learns ``indexing''. When the input is a longer text and the output includes multiple ssIDs for the concepts are presented in the input, the model is trained for ``recognition''. 

\subsection{Multi-level queries generated by LLMs}

Biomedical concepts are more challenging to recognize when the query is a passage compared to a sentence or span. By extracting shorter segments (e.g., sentences, phrases) from a passage, the model can better identify concepts that are difficult to capture when the query is a passage. Our framework, MA-COIR, is trained to process multiple levels of queries, enabling the integration of results from various query types derived from a passage into the final predictions. 

In this study, we employ an open-source LLM - llama-3-8b \cite{llama3modelcard}, to generate simplified queries from passages. For the CDR and HPO datasets, where concepts are associated with specific ``mentions'', the model generates concept names to serve as queries. Given that HOIP concepts are not consistently expressed as phrases, we use the model to transform passages into sentence-level claims and span-level concept names. 

Claims are prioritized over segmented sentences because they encapsulate the passage's meaning in a coherent and self-contained manner, facilitating comprehension and recognition. In contrast, segmented sentences often lack sufficient context, leading to ambiguity. Claims provide the necessary abstraction and semantic synthesis, aligning more effectively with downstream tasks that rely on conceptual understanding.

The concept name generation is performed under a 10-shot ICL setting. For a given passage in the test set, we randomly select 10 passage-concept pairs from the training set as demonstrations of the prompt.\footnote{Preliminary experiments using \(n\)-shot settings \((n=0, 1, 3, 5, 10)\) for LLM prompting on the HOIP dataset showed that the best results were achieved with a 10-shot setting.} Claim generation is done in a zero-shot setting due to the lack of annotated passage-claim pairs. Prompts we used are provided in Appendix Fig. \ref{fig:prompt_0}.


\subsection{Data augmentation}
\label{data-aug-detail}

After breaking down the passage into claims using an LLM on the HOIP dataset, we generate claim-ssID pairs from the training set for semi-supervised learning. This data construction follows a common weakly supervised NER approach, consisting of two steps:
\begin{itemize}
    \item Excerpt mining: Identify noun phrases and excerpts consisting of ``a noun phrase and a verb linked to that noun phrase'' using the dependency tree of a generated claim. We use spaCy \cite{spacy2} as the dependency parser.
    
    \item Labeling function: Represent each excerpt similarly to how a concept or query is represented, then compute the cosine similarity between the excerpt and annotated concepts from the passage. If any excerpt in the claim has a cosine similarity \(\geq 0.5\) to a gold concept, that concept is assigned to the claim.
\end{itemize}
Many matched excerpts only capture part of the meaning of the corresponding concept. Pairing the entire claim (which the excerpt appears) with the concept reduces noise compared to pairing the excerpt alone with the concept. An example of constructing a claim-concept instance is shown in Fig. \ref{fig:claim-concept}.

\begin{table}[t]
\small
    \centering
    \begin{tabular}{cccccc}
    \hline
         Split & Data & Passage & Claim & Concept & Mention \\
         \hline
         \multirow{3}{*}{Train} &CDR & 500& - & 1,328 & 2,672 \\
         &HPO & 182 & - & 416 & 926 \\
         &HOIP & 225 & 682 & 337 & - \\
         \hline
         \multirow{3}{*}{Test} & CDR & 500 & - &2,778 & 4,600 \\
         & HPO & 23 & - & 159 & 237 \\
         & HOIP & 37& 165 & 265 & - \\
         \hline
    \end{tabular}
    \caption{Statistics of instances.}
    \label{tab:setups1}
\end{table}

\section{Experiments}
\subsection{Datasets}
\label{datasets}

Target concepts in an ontology are expressed frequently either as mentions or not. The motivation for proposing MA-COIR is to apply a pragmatic approach for the latter. To evaluate the framework's effectiveness in both cases, we conduct experiments on the three datasets. 
\paragraph{CDR} The pair of the MeSH \footnote{\url{https://www.ncbi.nlm.nih.gov/mesh/}} and BC5CDR dataset \cite{DBLP:journals/biodb/LiSJSWLDMWL16}. 
The 2015 version of the MeSH vocabulary includes 258K terms and BC5CDR comprises 1,500 passages annotated with MeSH terms based on entity mentions. MeSH is not a formally defined ontology, evaluating performance on this scenario establishes a reference for the lower bound of ontological content.
\paragraph{HPO} The pair of Human Phenotype Ontology (HPO) \cite{10.1093/nar/gkad1005}\footnote{\url{https://hpo.jax.org/}} and HPO GSC+ dataset
published by \citet{https://doi.org/10.1155/2017/8565739}. The latest version of the HPO ontology includes over 19,000 concepts. The HPO GSC+ dataset comprises 228 PubMed abstracts and 1,933 mention annotations, each mention linked to a concept. 

\paragraph{HOIP} The pair of Homeostasis Imbalance Process (HOIP) ontology \cite{Yamagata2024} and HOIP dataset \cite{el-khettari-etal-2024-mention}.\footnote{\url{https://github.com/norikinishida/HOIP-dataset}} The ontology includes over 60,000 concepts related to homeostasis imbalance processes, of which 44,439 biological process concepts are target concepts. 

The dataset consists of 362 passages extracted from PubMed papers. Each passage is annotated with biological process concepts from the HOIP ontology. Mention annotations of concepts are not provided. Notably, a concept may be annotated based on its relevance to a process mentioned in the passage, even if the concept is not stated in the passage (this relevance may depend on the annotator's background knowledge). 

We conduct training with the original train/dev set, and evaluation with a refined test set containing only explicitly mentioned concepts.


\subsection{Comparison system} 
\paragraph{XR-Transformer.} Prior to MA-COIR, no supervised biomedical CR model directly generated a list of ontology concepts from free text. By treating concepts as labels, CR task can be naturally framed as an instance of extreme multi-label text classification (XMC), where a passage is assigned multiple relevant ontology terms. We adopt XR-Transformer \cite{zhang2021fast}, a state-of-the-art XMC model with top-tier performance across multiple public benchmarks, as a strong baseline.

\paragraph{kNN-searcher.}
Given the lack of existing approaches that do not use mentions for CR, we selected a straightforward baseline method: the top-k Nearest Neighbor (kNN) search, which can retrieve candidate concepts based on a given query.
As the way we represent a concept \(E_C\) that described in Section \ref{sec:concept-indexing}, we get the representation of the query \(E_Q\) by the \(TextEncoder\): 
\begin{equation}
  \label{eq:9}
  X_Q = TextEncoder(Q) \in \mathbb{R}^{l\times H}
\end{equation}
\begin{equation}
  \label{eq:10}
  E_Q = avg(X_Q) \in \mathbb{R}^H
\end{equation}
where \(l\) is the token length of the query, and \(H\) is the dimension of a token’s embedding.

With \(E_Q\) and representations of all concepts \(\{E_{C_1}, ..., E_{C_n}\}\) as input vectors, we implemented Faiss \cite{douze2024faiss} for a fast vector search of \(E_Q\) among large-scale concept spaces, by calculated similarity based on Euclidean distance. The kNN-searcher may return a candidate even if its distance from the query is large, when no other concepts closer to the query exceed the distance of the candidate. To mitigate false positives, we classify retrieved concepts with a similarity score \(< 0.6\) as non-predictions.

Additionally, we conduct a comparative analysis of our approach against \cite{shlyk-etal-2024-real} and \cite{el-khettari-etal-2024-mention} under a controlled setup. Details are described in Appendix~\ref{sec:more-res}


\subsection{Setups}
\label{setups}
\begin{table*}[t]
\small
\centering
\begin{tabular}{c|c|c|ccc|ccc|ccc}
\hline
\multirow{2}{*}{Dataset}&\multirow{2}{*}{k}& \multirow{2}{*}{Query} & \multicolumn{3}{c|}{MA-COIR}& \multicolumn{3}{c|}{XR-Transformer}& \multicolumn{3}{c}{kNN-searcher}\\

& & & Pre& Rec& F1& Pre& Rec& F1& Pre& Rec& F1\\
\hline
\multirow{9}{*}{CDR}&\multirow{3}{*}{1} & Passage & 51.0 & \textbf{44.6} & \textcolor{red}{\textbf{47.6}} & \textbf{79.6} & 11.6 & 20.3 & 13.3 & 0.1 & 0.1 \\
 && Mention & 67.2 & 72.0 & 69.5 & 67.1 & 71.4 & 69.1 & \textbf{75.5} & \textbf{82.5} & \textcolor{red}{\textbf{78.9}} \\
 && Concept & 57.2 & 41.2 & 47.9 & 57.2 & 41.5 & 48.1 & \textbf{63.5} & \textbf{48.2} & \textcolor{red}{\textbf{54.8}} \\
 \cline{2-12}
&\multirow{3}{*}{5} & Passage & 36.5 & \textbf{49.6} & \textbf{42.0} & \textbf{45.3} & 33.1 & 38.3 & 12.5 & 0.1 & 0.2 \\
 && Mention & 17.1 & 74.8 & 27.9 & 13.8 & 73.6 & 23.3 & \textbf{18.9} & \textbf{92.0} & \textbf{31.3} \\
 && Concept & 15.2 & 44.2 & 22.6 & 12.4 & 44.4 & 19.4 & \textbf{16.5} & \textbf{56.1} & \textbf{25.5} \\
 \cline{2-12}
&\multirow{3}{*}{10}& Passage & \textbf{29.9} & \textbf{52.0} & \textbf{37.9} & 26.7 & 39.0 & 31.7 & 10.5 & 0.1 & 0.2 \\
 && Mention & 9.2 & 75.5 & 16.4 & 7.1 & 74.1 & 13.0 & \textbf{11.4} & \textbf{93.1} & \textbf{20.3} \\
 && Concept & 8.3 & 45.4 & 14.0 & 6.4 & 44.8 & 11.2 & \textbf{9.9} & \textbf{57.3} & \textbf{16.9} \\
\hline

\multirow{9}{*}{HPO}&\multirow{3}{*}{1}& Passage & 67.7 & \textbf{53.8} & \textcolor{red}{\textbf{60.0}} & \textbf{91.3} & 13.5 & 23.5 & 33.3 & 0.6 & 1.3 \\
 && Mention & 85.6 & 80.1 & 82.8 & \textbf{88.1} & \textbf{85.3} & \textcolor{red}{\textbf{86.6}} & 70.7 & 71.2 & 70.9 \\
 && Concept & \textbf{65.9} & 57.1 & 61.2 & 65.2 & \textbf{57.7} & \textcolor{red}{\textbf{61.2}} & 58.5 & 50.6 & 54.3 \\
 \cline{2-12}
&\multirow{3}{*}{5} & Passage & 60.8 & \textbf{57.7} & \textbf{59.2} & \textbf{61.7} & 45.5 & 52.4 & 11.1 & 0.6 & 1.2 \\
 && Mention & 21.2 & 84.0 & 33.8 & 19.2 & \textbf{87.8} & 31.5 & \textbf{21.3} & \textbf{87.8} & \textbf{34.3} \\
 && Concept & \textbf{18.5} & \textbf{66.7} & \textbf{29.0} & 15.4 & 66.0 & 25.0 & 18.1 & \textbf{66.7} & 28.4 \\
 \cline{2-12}
&\multirow{3}{*}{10}& Passage & \textbf{54.1} & 59.6 & \textbf{56.7} & 43.9 & \textbf{64.7} & 52.3 & 7.7 & 0.6 & 1.2 \\
 && Mention & 12.4 & 87.2 & 21.7 & 9.9 & 87.8 & 17.7 & \textbf{13.9} & \textbf{89.1} & \textbf{24.0} \\
 && Concept & \textbf{11.0} & \textbf{73.7} & \textbf{19.2} & 8.2 & 67.9 & 14.6 & \textbf{11.0} & 67.9 & 18.9 \\
\hline
\end{tabular}
  \caption{\label{tab:results-cdr-hpo}
    Results of the top-\(k\) generated sequences by MA-COIR and the top-\(k\) retrieved concepts by the XR-transformer and kNN-searcher on the CDR and the HPO.  ``mention'' are gold annotated mentions of a passage. ``concept'' are generated concepts by the LLM given a passage. Red values indicate the highest F1 score achieved for each query type on a given dataset.
  }
\end{table*}

We trained MA-COIR and XR-Transformer using passage-, name-, and synonym-ssID pairs for all three datasets. When annotated mentions or generated claims were available, the model was trained with mention- and claim-ssID pairs. The models trained with synthetic claim-ssID pairs is referred to as \textbf{MA-COIR-a} and \textbf{XR-Transformer-a}. For checkpoint selection, we used only passage-ssID pairs from the development set.
Evaluation involved testing the model with various types of queries, including passages, gold mentions (for CDR and HPO), generated claims (only for HOIP), and generated concept names. The statistics for the instances are provided in Table~\ref{tab:setups1}. Hyperparameters are listed in Appendix \ref{app:hyperp}.

\subsection{Evaluation metrics} 



We evaluate all models using precision (Pre), recall (Rec), and micro F1-score (F1), measured across different query levels. For MA-COIR, we use beam search to generate top-\(k\) concept sequences per query. Each sequence is segmented into ssID-like spans using semicolons as delimiters. Spans not matching any defined ssID are discarded. All valid spans across \(k\) sequences are then merged and deduplicated to form the final prediction set. When multiple queries are derived from a single passage, their predictions are aggregated and compared against the gold annotations for that passage.

To ensure a fair comparison, passage-level input for the kNN-searcher is the same full-text passage used by MA-COIR, rather than shorter fragments obtained via "excerpt mining" we described in Section~\ref{data-aug-detail}.

\begin{table*}
\small
\centering
\begin{tabular}{c|c|ccc|ccc|ccc|ccc}
\hline
\multirow{2}{*}{k}& \multirow{2}{*}{Query} & \multicolumn{3}{c|}{MA-COIR}&\multicolumn{3}{c|}{MA-COIR-a}& \multicolumn{3}{c|}{XR-Transformer-a}& \multicolumn{3}{c}{kNN-searcher}\\

& & Pre& Rec& F1& Pre& Rec& F1& Pre& Rec& F1& Pre& Rec& F1\\
\hline
\multirow{3}{*}{1}& Passage & 11.1 & 25.0 & 15.4 & 13.0 & \textbf{27.3} & 17.6 & \textbf{32.4} & 13.6 & \textbf{19.2} & 6.7 & 2.3 & 3.4 \\
 & Claim & 8.2 & 21.6 & 11.9 & 14.1 & \textbf{30.7} & 19.3 & \textbf{19.8} & 28.4 & \textcolor{red}{\textbf{23.4}} & 6.7 & 8.0 & 7.3 \\
 & Concept & 18.2 & 46.6 & 26.2 & \textbf{18.5} & \textbf{48.9} & \textcolor{red}{\textbf{26.8}} & 17.8 & 45.5 & 25.6 & 13.0 & 35.2 & 19.0 \\
\hline
\multirow{3}{*}{5} & Passage & 8.6 & 34.1 & 13.8 & 11.0 & \textbf{39.8} & 17.2 & \textbf{14.6} & 30.7 & \textcolor{red}{\textbf{19.8}} & 2.1 & 3.4 & 2.6 \\
 & Claim & 6.0 & 45.5 & 10.7 & \textbf{7.4} & \textbf{47.7} & \textbf{12.8} & 6.5 & 45.5 & 11.4 & 3.8 & 17.0 & 6.3 \\
 & Concept & 6.4 & 64.8 & 11.6 & \textbf{6.7} & \textbf{68.2} & \textbf{12.1} & 5.5 & 64.8 & 10.1 & 5.0 & 56.8 & 9.1 \\
\hline
\multirow{3}{*}{10}& Passage & 7.2 & 36.4 & 12.0 & 9.8 &\textbf{45.5} & \textbf{16.2} & \textbf{10.0} & \textbf{42.0} & 16.2 & 2.4 & 6.8 & 3.6 \\
 & Claim & 4.7 & 54.5 & 8.7 & \textbf{5.9} & \textbf{59.1} & \textbf{10.7} & 4.2 & 55.7 & 7.8 & 2.6 & 17.0 & 4.4 \\
 & Concept & 3.9 & 69.3 & 7.4 & \textbf{4.4} & \textbf{78.4} & \textbf{8.4} & 3.0 & 69.3 & 5.7 & 3.3 & 62.5 & 6.2 \\
\hline
\end{tabular}
  \caption{\label{tab:results-hoip}
    Results of the top-\(k\) generated sequences by MA-COIR and the top-\(k\) retrieved concepts by the XR-Transformer and kNN-searcher on the HOIP dataset. ``claim'' and ``concept'' refer to generated claims and concepts, produced by the LLM given a passage. Red values indicate the highest F1 score achieved for each query type.
  }
\end{table*}

\section{Results}

Tables~\ref{tab:results-cdr-hpo} and \ref{tab:results-hoip} summarize model performance across three biomedical concepts. On both CDR and HPO, MA-COIR consistently achieves the best F1 scores with passage-level inputs (47.6 and 60.0, respectively), while kNN-searcher and XR-Transformer perform best with span-level inputs. In the more challenging HoIP setting, MA-COIR-a and XR-Transformer-a outperform kNN-searcher, with XR-Transformer-a achieving the highest F1 for passage- and claim-level inputs ((19.8 and 23.4), and MA-COIR leading in the span-level setting (26.8). We analyze results from three complementary perspectives: concept type, input granularity, and real-world applicability.

\paragraph{Concept Type.}
The three datasets involve concept spaces of increasing complexity—from chemical and drug names (CDR), to phenotype abnormalities (HPO), and finally to abstract homeostasis imbalance processes (HoIP).

In CDR, most gold concepts are explicitly mentioned in text or have close surface-level synonyms, making the kNN-searcher highly effective. However, HPO concepts such as ``Abnormality of body height'' or ``Abnormal platelet morphology'' are semantically richer and less likely to appear verbatim. Here, supervised models like MA-COIR and XR-Transformer gain a clear edge by leveraging learned task-specific information.

HoIP presents the greatest challenge: many target concepts are abstract, fine-grained, and rarely expressed via identifiable mentions, challenging to recognize even for experts (e.g., ``dysregulation of matrix metalloproteinase secretion''). In addition, HoIP lacks mention-ssID training pairs, limiting supervised grounding.\footnote{A study examining the impact of mention information on MA-COIR, conducted on CDR, revealed a significant difference with and without mention-ssID pairs as training data, as detailed in Appendix \ref{app:train-abl}.} As a result, all models struggle, but the gap between supervised and unsupervised methods widens. This underscores a key insight: concept complexity and the mentioned way are critical determinants of method suitability.

\paragraph{Input Granularity.}
MA-COIR excels with passage-level inputs, outperforming XR-Transformer by large margins on CDR (47.6 vs. 38.3) and HPO (60.0 vs. 52.4), and achieving stronger recall on HoIP. The kNN-searcher, by contrast, underperforms in this setting due to poor alignment between full passages and span-based embeddings.

At the span-level, performance varies: MA-COIR outperforms XR-Transformer when given gold mentions on CDR, but lags slightly on HPO. When using concept names generated by LLMs, MA-COIR matches or exceeds XR-Transformer. This reflects the robustness of MA-COIR to input variation and highlights a key practical strength: in real applications, gold mentions are unavailable, and LLM-generated spans often differ in granularity from ontology entries, making retrieval harder. MA-COIR's adaptability makes it better suited for such realistic, mention-free scenarios.

\paragraph{Practical Considerations.}
On CDR and HPO, MA-COIR demonstrates strong and consistent performance, proving its effectiveness for real-world biomedical CR. On HoIP, XR-Transformer-a achieves slightly higher F1 than MA-COIR-a (19.8 vs. 17.6). This is largely due to the dataset’s statistics: each passage contains, on average, 7.2 gold concepts. XR-Transformer-a’s fixed-$k$ retrieval (with $k=5$) benefits from limiting false positives, whereas MA-COIR-a uses beam search to generate unbounded concept sequences, trading off precision for recall.
In practice, however, concept density varies across documents, and setting an optimal $k$ is non-trivial, limiting the robustness of fixed-$k$ methods like XR-Transformer.

On span-level CDR tasks, MA-COIR and XR-Transformer perform comparably, but both fall short of kNN-searcher when provided with gold mentions. On HPO, kNN-searcher is only competitive when given gold mentions and big $k$ values (e.g., $k=5$ or $10$).
Further analysis (Appendix~\ref{app:recall0}) reveals that MA-COIR struggles to recognize unseen concepts lacking training exposure—an issue shared with XR-Transformer. In contrast, kNN-searcher remains unaffected. Nonetheless, we believe this limitation can be mitigated via data synthesis strategies: our preliminary experiments confirm the feasibility of using synthetic samples to improve MA-COIR’s generalization.

\paragraph{Summary.}
MA-COIR delivers strong performance across diverse concept types and input settings. While training data coverage remains a limitation, this can be addressed with scalable augmentation techniques. Given its flexibility, robustness to input variation, and effectiveness even without gold mentions, MA-COIR offers a practical and reliable solution for biomedical CR.
\section{Analysis}

\subsection{Effectiveness of ssID}

To verify the effectiveness of ssID, we compared it with other types of indexes can be used for the recognition on the HOIP. 
\begin{itemize}
    \item Random ID: Randomly assign a number to each concept as an index. The index ranges from 0 to the number of all ontology concepts.
    \item Ontology ID: The unique ID of each concept in the ontology is used as the index. Like ``HOIP\_0004832'' is the ontology ID of ``TNF signaling'', and the index for generation.
    \item ssID (name): As described in Section \ref{sec:concept-indexing}.
    \item ssID (+hypernyms): The indexes are based on constructing a label tree using the concatenation of the representation of a name of each concept, and the average of the representations of its hypernyms. The hypernymy and hyponymy relations is known from the ontology. 
    Let \(U_C\) denote a set of concepts that are hypernyms of concept \(C\) defined in the ontology. The representation of the concept \(C\) used for label tree construction changed from eq. \ref{eq:2} to eq. \ref{eq:4}.
    \begin{equation}
    \label{eq:11}
    E_{{U_C}_i} = avg(X_{{U_C}_i}) \in \mathbb{R}^H
    \end{equation}
    \begin{equation}
    \label{eq:12}
    E_C = [avg(X_C): avg(E_{U_C})]
    \end{equation}
where ``\(:\)'' is the concatenation operation, \(H\) is the dimension of a token’s embedding, \(E_C \in \mathbb{R}^{2 \times H}\).

\end{itemize}

\begin{table}
\small
  \centering
  \begin{tabular}{lcccc}
  \hline
Index type& Pre& Rec& F1\\
  \hline
Random ID& 7.8& 31.8& 12.5\\
Ontology ID& 6.7& \textbf{47.7}& 11.8\\
ssID (name)& \textbf{11.1}& 25.0& \textbf{15.4}\\
ssID (+hypernyms)& 9.7& 20.5& 13.1\\
  \hline
  \end{tabular}
  \caption{
  \label{tab:ssid_hoip_e}
    Results of the top-1 generated sequence using various index types with the passage queries on the HOIP dataset by MA-COIR.
  }
  
\end{table}

\begin{figure}[t]
\centering
  \includegraphics[width=0.85\linewidth]{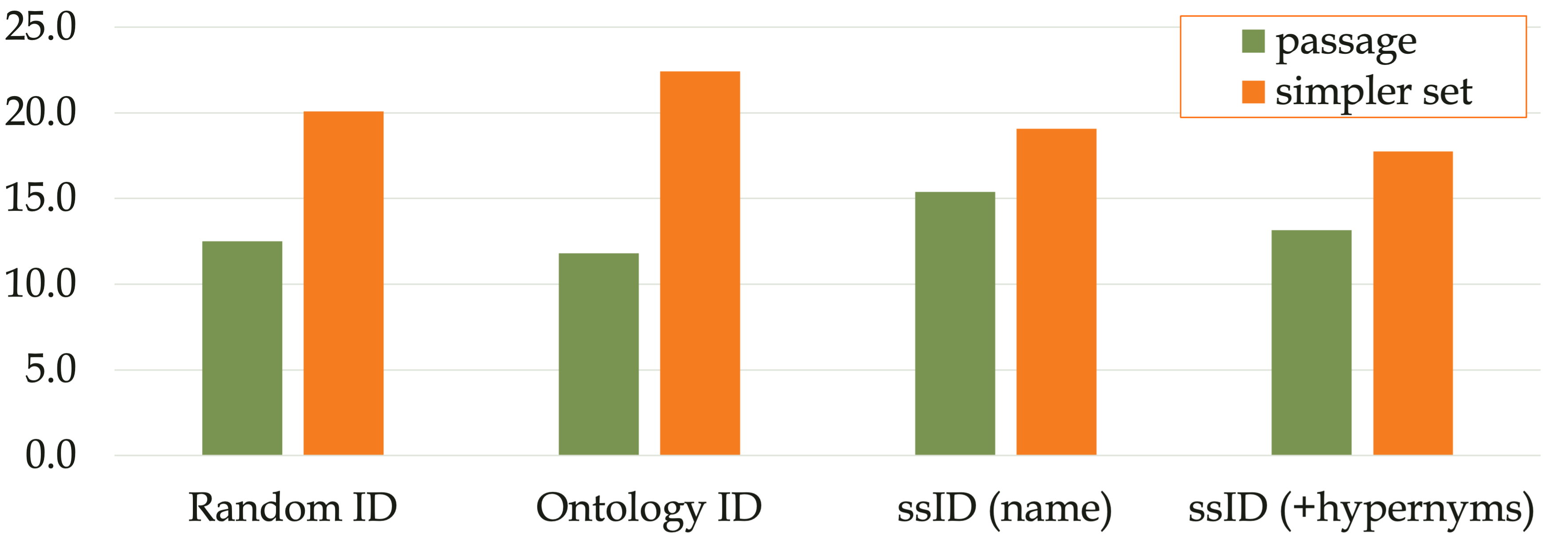}
  \caption {F1 scores by MA-COIR between complex query (passage) and the average of the simpler set of queries (claim/concept) from top-1 generated sequence using different indexes on the HOIP.}
  \label{fig:ssid-f1}
\end{figure}

The experimental results are summarized in Table \ref{tab:ssid_hoip_e}. Both Random ID and Ontology ID performed well on span-level queries, providing higher recall compared to ssIDs. On the other hand, using ssID (name) achieved the highest precision and F1 scores for passage-level queries. As shown in Fig. \ref{fig:ssid-f1}, ssID-based indexing demonstrates robustness across both complex and simple queries, whereas Random ID and Ontology ID perform optimally only on shorter queries. In the absence of tools to retrieve non-passage level information, ssID is clearly the superior choice.

\subsection{Effectiveness of data augmentation} 

\begin{table}
\small
\centering
\begin{tabular}{llccc}
\hline
Query & Pre& Rec& F1 \\
\hline
passage& 13.0 & 27.3 & 17.6\\
\hspace{0.2cm} + claim & 12.5 & 45.5& 19.7\\
\hspace{0.4cm} + concept & 12.3& \textbf{64.8} & 20.7\\
\hspace{0.2cm} + concept & \textbf{14.7} & 61.4 & \textbf{23.7}\\
\hline
\end{tabular}
  \caption{\label{HOIP-res-1}
    Results of the top-1 generated sequence by MA-COIR-a on HOIP. 
  }
\end{table}

The results for the MA-COIR-a are presented in Table \ref{tab:results-hoip}. Incorporating claim-ssID pairs, as described in Section \ref{data-aug-detail}, leads to improvements across all metrics for all query types. F1 scores for claim-queries increase by 4.6 points compared to MA-COIR. Across all query types, the improvement in recall exceeds that in precision, indicating that the added data is both accurate (with minimal noise, which helps maintain precision) and diverse, benefiting all query types.

\subsection{Combination of different-level queries}

The results of combining predictions of various types of queries are presented in Table \ref{HOIP-res-1}. While the accuracy of decomposing full passages into shorter units is low, MA-COIR captures additional concepts that are difficult to detect from full-length inputs alone. The predictions from different query levels exhibit partial but non-trivial overlap, revealing their complementary strengths.

Each query type offers distinct advantages.
Aggregating predictions across all levels yields substantial gains. Recall improves significantly from \((27.3\rightarrow45.5\rightarrow64.8)\) when integrating all three, underscoring the value of multi-level querying.

\section{Conclusion}
We present the MA-COIR framework, a flexible and implementable 
solution for recognizing both simple and complex biomedical concepts explicitly or implicitly appeared in scientific texts, without requiring specific mention information. The framework meets the needs of domain experts, as demonstrated by experiments on three vocabulary/ontology-dataset pairs. We introduce efficient methods for obtaining queries at various levels and data augmentation using an LLM and proving their efficacy in low-resource scenarios. 
MA-COIR’s adaptability to multi-level queries enhances its practical utility. We further provide an in-depth analysis of biomedical concept recognition and potential directions for future improvement.

\section*{Limitations}

Although we would like MA-COIR to generate ssIDs for unseen concepts based on semantic similarities with seen concepts, results indicate that it lacks this capability. This restricts the model's applicability to the available dataset. Given that the annotated dataset contains significantly fewer concepts than the full ontology, further framework refinement is needed to allow comprehensive processing across different input levels and consistent mapping of all ontology concepts and their indexes.

It is essential to develop validation datasets that align with the needs of domain experts. In the HPO and HOIP test sets, the low proportion of unseen concepts limits the evaluation of the model's generalization to out-of-dataset concepts. Without observing MA-COIR's performance decline on the CDR dataset, this limitation might have gone unrecognized.

Last but not least, the performance of MA-COIR also depends on query quality. There is a substantial gap between results for concept names generated by an LLM and those derived from gold annotated mentions. Although we have not fully explored LLM-based query generation, it is unrealistic to expect consistent query quality across specialized biomedical domains. Thus, it is critical to both improve the model's robustness to lower-quality queries and identify ways to generate high-quality queries.


\bibliography{acl_latex}

\appendix
\section{Appendix}
\label{sec:appendix}

\subsection{Hyperparameters}
\label{app:hyperp}

The BART-based language model (facebook/bart-large) used in MA-COIR for recognition is trained with hyperparameters listed in the Table \ref{tab:hyp}.

The hyperparameters of the K-Means clustering algorithm used for hierarchical clustering process, are \(g\) and \(m\), while \(g\) is the maximum number of the elements covered by a node when we can stop further dividing the node into smaller clusters. \(m\) is the number of clusters when we divide the elements in a node. For example, when \(g=10, m=10\), if there are 9 elements in the current node, we do not divide the elements in this node by clustering; if there are 18 elements in the current node, we will do a clustering for these elements, so that these elements will be categorized into \(m=10\) clusters. In this work, we set \(g=10, m=10\).

For the training of XR-Transformer, we implement the model with the library pecos\footnote{https://pypi.org/project/libpecos/}, setting the hyperparameters provided by the authors, as those have already been tuned. The architecture of the Transformers model we used in the experiments is BERT.

\begin{table}[t]
\small
    \centering
    \begin{tabular}{c|c}
    \hline
    Item & Value \\
    \hline
         model\_card &  facebook/bart-large\\
         learning\_rate& 1e-5\\
         num\_epoch& 50\\
         batch\_size& 4\\
         max\_length\_of\_tokens & 1024 \\
    \hline
    \end{tabular}
    \caption{Hyperparameters of the recognizer.}
    \label{tab:hyp}
\end{table}

\subsection{LLM Application}

\begin{figure*}[t]
    \centering
    \includegraphics[width=1.0\linewidth]{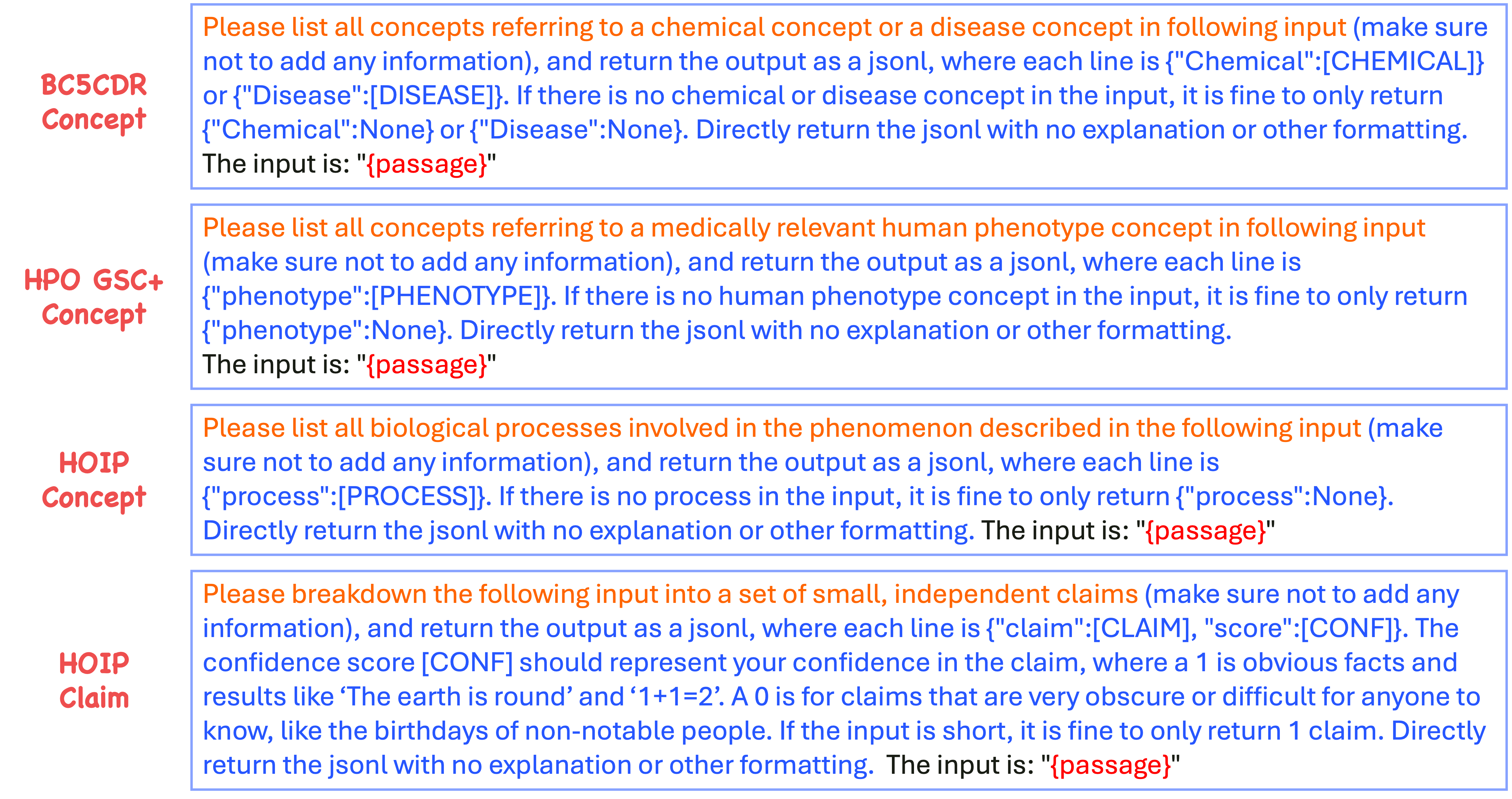}
    \caption{\label{fig:prompt_0}Prompt template for generating concept names / claims for passage. A prompt consists of \textcolor{orange}{task instruction}, \textcolor{blue}{output format instruction}, several demonstrations and the \textcolor{red}{query}. }
\end{figure*}

We applied a large language model llama-3-8b for query generation. For all concept generation tasks, the prompt consists of ``instruction'', ``n demonstrations'' under the n-shot setting, and the passage. The prompts we used for concept name generation on CDR, HPO and HOIP are shown in Fig. \ref{fig:prompt_0}.
For claim generation, the prompt template we used for a passage on HOIP is shown in Fig. \ref{fig:prompt_0}. The generation is conducted in a zero-shot scenario cause there is no annotated data for passage-claim pairs.

\subsection{Performance on seen and unseen concepts}
\label{app:recall0}
Upon examining MA-COIR's performance on both seen (concepts appeared in the training set) and unseen concepts (concepts only appeared in the test set), we found that the performance gap between it and the kNN-searcher is primarily due to its inability to recognize unseen concepts. As presented in the Table \ref{tab:unseen}, when we evaluated the model on unseen concepts, MA-COIR achieved a recall of nearly 0.0 on both the CDR and the HPO. 
\begin{table}
\small
    \centering
    \begin{tabular}{cccccc}
    \hline
         & & \multicolumn{2}{c}{CDR}& \multicolumn{2}{c}{HPO}\\
k& Query& Seen& Unseen& Seen& Unseen\\
\hline
& passage& 57.2& 0.3& 60.0& 0.0\\
1& mention& 92.4& 0.0& 89.3& 0.0\\
& concept& 52.9& 0.0& 63.6& 0.0\\
\hline
& passage& 63.6& 0.3& 64.3& 0.0\\
5& mention& 95.2& 2.9& 92.1& 12.5\\
& concept& 56.3& 1.5& 74.3& 0.0\\
\hline
& passage& 66.6& 0.4& 66.4& 0.0\\
10& mention& 95.8& 4.0& 95.0& 18.8\\
& concept& 57.7& 2.2& 80.7& 12.5 \\
\hline
    \end{tabular}
    \caption{Recalls on the seen and unseen concepts of the top-\(k\) generated sequences by MA-COIR.}
    \label{tab:unseen}
\end{table}

\subsection{Training data for ``Indexing'' capability of the recognizer}
\label{app:train-abl}

\begin{table}
\small
  \centering
  \begin{tabular}{lcccc}
  \hline
   Data & Query & Pre & Rec & F1\\ 
  \hline
    \multirow{3}{*}{All}& passage&51.0& 44.6& 47.6\\ 
    & mention&67.2&72.0&69.5\\ 
    & concept&57.2&41.2&47.9\\ 
  \hline
     \multirow{3}{*}{- mention}& passage& 36.1& 30.5& 33.1\\ 
    & mention& 39.5& 42.8& 41.1\\ 
    & concept& 32.4& 22.3& 26.4\\ 
  \hline
    \multirow{3}{*}{- synonym}& passage& 48.2& 42.3& 45.0\\ 
    & mention& 67.4& 72.0& 69.6\\ 
    & concept& 58.2& 41.4& 48.3\\ 
    \hline
    - mention& passage& 36.0& 30.5& 33.0\\ 
    \hspace{0.2cm}- synonym& mention& 41.9& 44.8& 43.3\\ 
    & concept& 37.6& 24.8& 29.9\\
  \hline
  \end{tabular}
  \caption{\label{train-data-abl}
    Results on CDR with different training data. ``All'' contains passage-ssIDs pairs, name-ssID pairs, synonym-ssID pairs and mention-ssID pairs constructed from the original training set.
  }
\end{table}

The indexing capability of the model refers to the model's ability to generate the correct ssID for the query when it is a span. On datasets labelled with mentions, in addition to the canonical names and synonyms of a concept in the ontology that can be used to train model indexing capabilities, mentions are also very effective data. We conducted an ablation study on the CDR dataset to confirm the impact of synonym- and mention-ssID information on the model's ability to recognize concepts. The results can be seen in Table \ref{train-data-abl}.

After removing the mention-ssID data, the model's performance dropped significantly; removing the synonym-ssID data, the performance on the passage-level query dropped less and even improved on the span-level query. This illustrates that the way a concept is expressed within a particular application (passage) is important for capturing the relationship between the concept and the ssID. Not only the indexing capability are influenced by removing mention data, but also the recognition on the passage query (\(\downarrow 14.5\) F1 score). 
The slight improvement after removing synonym-ssID pairs indicates how different the common expressions written in scientific papers and the technical terms of a concept are. Using synonyms to enrich concept information makes the query and a concept further apart in representation.

\subsection{More comparisons}
\label{sec:more-res}

\begin{table}[t]
\small
    \centering
    \begin{tabular}{ccccc}
    \hline
         Dataset & Method & Pre & Rec & F1\\
         \hline
         \multirow{4}{*}{HPO} & REAL-1st hit & 40.0 &49.0&44.0\\
         & REAL-GPT3.5 & \textbf{68.0} &48.0&56.0\\
         & kNN-searcher & 58.5& 50.6& 54.3\\
         & MA-COIR & 63.4&\textbf{54.5}&\textbf{58.6}\\
         \hline
         \multirow{3}{*}{HOIP-o} & \cite{el-khettari-etal-2024-mention} & \textbf{43.1} &11.8&18.6\\
         & kNN-searcher & 42.0& 13.9& 20.9\\
         & MA-COIR &23.7 &\textbf{19.6} &\textbf{21.5}\\
         \hline
    \end{tabular}
    \caption{\label{tab:hpo-further}
    Comparison between our methods and previous works. ``HOIP-o'' refers to the original test set. 
    }
\end{table}

Our framework operates under different setups compared to previous studies that were validated on the same dataset. We provide results using a more comparable setting to ensure fair evaluation (see Table \ref{tab:hpo-further}). 

For HPO dataset, REAL \cite{shlyk-etal-2024-real} reports results for two approaches: for an LLM generated mention, selecting the top-1 candidate from three candidates provided to GPT-3.5 (REAL-GPT3.5) or taking the top-1 concept retrieved by kNN searching (REAL-1st hit). 
We report the results by MA-COIR trained without mention-ssID pairs and the kNN-searcher we implemented using concept queries with \(k=1\).

For HOIP dataset, \citet{el-khettari-etal-2024-mention} report the results of a similarity-based kNN search for concepts generated by llama-3-8b in its few-shot setting (ICL-Llama). After retrieval, they filtered out out-of-dataset predictions. We replicated their approach by using their generated concepts as queries and applying the same filter with kNN-searcher and setting \(k=1\).

From the results of REAL-1st hit and kNN-searcher on HPO (F1: 44.0/54.3), as well as kNN-searcher on concepts from ICL-Llama and our generated concepts (F1: 18.6/20.9) on HOIP-o, we can infer that the quality of our generated concepts and the representation of concepts/queries is consistent with previous methods.

The removal of out-of-dataset concepts significantly reduced false positives in similarity-based methods, improving precision to over 40.0 on the HOIP-o. In contrast, MA-COIR does not predict concepts never appeared in the training phase, such post-processing does not provide benefits.

Overall, our supervised recognizer, MA-COIR, outperforms unsupervised LLM-based solutions like REAL-GPT3.5 and ICL-Llama.

\end{document}